\documentclass[letterpaper, 10 pt, conference]{ieeeconf}  % Comment this line out if you need a4paper

\IEEEoverridecommandlockouts                              % This command is only needed if 
\DeclareTextSymbolDefault{\dh}{T1}
\usepackage{hyperref}
\usepackage{dblfloatfix}
\usepackage{graphicx}
\usepackage{amsmath} % assumes amsmath package installed
\usepackage{amssymb}  % assumes amsmath package installed
\usepackage{bm}  % assumes amsmath package installed
\usepackage{tikz}
\usepackage{pgfplots}
\usepgfplotslibrary{fillbetween}
\usepackage{subcaption}
\usepackage{algorithm}
\usepackage{xcolor}
\usepackage{tabularx}
\usepackage{multicol}
\usepackage{multirow}
\usepackage{booktabs}
\newcolumntype{Y}{>{\centering\arraybackslash}X}
\newcolumntype{Z}{>{\centering\arraybackslash}p{0.85cm}}
\newcolumntype{K}{>{\centering\arraybackslash}p{0.6cm}}
\newcolumntype{M}[1]{>{\centering\arraybackslash}m{#1}}
\usepackage{array}
\usepackage[dvipsnames]{xcolor}
\usepackage{balance}
\usepackage{algorithmic}

\newcommand{\rstruct}[1]{struct~\ref{#1}}

\newcommand{\reffig}[1]{Fig. \ref{#1}}
\newcommand{\refeq}[1]{Eq. (\ref{#1})}
\newcommand{\refalg}[1]{Algorithm \ref{#1}}

\newcommand{\refsec}[1]{Section \ref{#1}}
\newcommand{\refsecs}[1]{Sec. \ref{#1}}

\newcommand{\algo}[0]{CASSR}

\newcommand{\vc}[1]{\mathbf{#1}} 					% Vector symbol
\def\@onedot{\ifx\@let@token.\else.\null\fi\xspace}

\newcommand{\degree}{\ensuremath{^\circ}}				% define the degree symbol
\newcommand{\Expect}{{\rm I\kern-.3em E}}				% expectation
\newcommand{\astr}[0]{\ensuremath{A^*} }
\newcommand{\astrns}[0]{\ensuremath{A^*}}

\title{\LARGE \bf
CASSR: Continuous A-Star Search through Reachability for real time footstep planning
}
\pgfplotsset{compat=1.18}

\author{Jiayi Wang\thanks{Jiayi Wang is with State Key Laboratory of General Artificial Intelligence, BIGAI, Beijing 100080, China, wangjiayi@bigai.ai} and Steve Tonneau\thanks{Steve Tonneau is with the University of Edinburgh, IPAB, Edinburgh EH8 9YL, U.K. stonneau@ed.ac.uk}} % <-this % stops a space}

\begin{document}

\maketitle

%%%%%%%%%%%%%%%%%%%%%%%%%%%%%%%%%%%%%%%%%%%%%%%%%%%%%%%%%%%%%%%%%%%%%%%%%%%%%%%%
\begin{abstract}

Footstep planning involves a challenging combinatorial search. Traditional A* approaches require discretising reachability constraints, while Mixed-Integer Programming (MIP) supports continuous formulations but quickly becomes intractable, especially when rotations are included. 
We present \algo{}, a novel framework that recursively propagates convex, continuous formulations of a robot’s kinematic constraints within an A* search. Combined with a new cost-to-go heuristic based on the EPA algorithm, \algo{} efficiently plans contact sequences of up to 30 footsteps in under 125 ms. Experiments on biped locomotion tasks demonstrate that \algo{} outperforms traditional discretised A* by up to a factor of 100, while also surpassing a commercial MIP solver. These results show that \algo{} enables fast, reliable, and real-time footstep planning for biped robots. Video: \url{http://youtu.be/reDGK-VXg9k}

\end{abstract}

%%%%%%%%%%%%%%%%%%%%%%%%%%%%%%%%%%%%%%%%%%%%%%%%%%%%%%%%%%%%%%%%%%%%%%%%%%%%%%%%
\section{INTRODUCTION}
Legged robots act by making and breaking contact with their environment. Planning the sequence of contacts required to achieve a task (e.g. walk to reach a point in space) is a discrete problem of exponential complexity\cite{Bretl2006MotionPO}. It can be practically tackled using graph-based techniques, where each node corresponds to a set of active contacts. These techniques include sampling based approaches~\cite{escande2013planning, tonneau2018efficient}, Mixed Integer Programme solvers (MIP) \cite{yunt2006trajectory, deits2014footstep}, or  \astrns variants~\cite{chestnuttkuffner,griffin2019footstep,kumagai2020multi}.

These approaches require modelling 1-step reachability constraints, i.e. characterising  which contact states can be reached in one contact transition from the current state.  These constraints are non-linear, thus hard to handle simultaneously with the combinatorics. They are often approximated, thus efficiency is defined both in terms of success rate and computational performance. Convex linear approximations of reachability \cite{wieber} provide a locally continuous representation of the feasible space and enable interactive resolution through sampling-based and MIP solvers.

However, continuous models of 1-step reachability are not compatible with \astr search. Existing approaches instead use a discrete representation of the reachable space~\cite{chestnuttkuffner,griffin2019footstep,kumagai2020multi}, which reduces the number of solutions that can be found and aggravates the combinatorial complexity of the problem. Still, \astr presents several advantages: it is deterministic and encodes optimality by design; it is simpler to implement than a MIP solver; it has a lower complexity on shortest-path problems than the other approaches. These advantages motivate the search for \textbf{a continuous formulation of reachability constraints compatible with \astrns}, which would enable a more exhaustive and computationally efficient search.

In this work, we propose to reformulate \astr not as the search for a sequence of contact locations, but rather as the search for a sequence of contact \textbf{surfaces}. Our method uses a recursive formulation of the reachability constraints \cite{nas,bansal2017hamilton,althoff2021set} to expand the search graph, which guarantees that the contact surface sequence can be transformed into a feasible sequence of contact locations through the resolution of a simple Quadratic Program (QP). We call our algorithm \algo{}: Continuous A-Star Search through Reachability for real time
footstep planning.  Our results on biped locomotion problems experimentally validate the computational superiority of \astr algorithms over MIP approaches. Quantitative analysis of the number of nodes explored and the computational performance show that \algo{} outperforms traditional \astr by a factor up to a 100.

Our main contributions are:

\begin{itemize}
       \item The first \astr solver with a continuous reachability representation, for real-time planning over long horizons.
    \item A novel cost-to-go heuristic using the EPA algorithm~\cite{vandenBergen2001} to compute polytope–target distances, efficiently approximating rotation distances.
    \item A novel, robust safety cost that maximises the minimum distance to contact surface borders.
   % \item An experimental framework demonstrating that our method explores fewer nodes and achieves substantial computational speed-ups compared to both discretised \astr and a commercial MIP solver, enabling real-time planning on challenging locomotion problems.
\end{itemize}

%Our experimental framework is available in an open-source form, anonymised for the submission, but to be released upon acceptance.

\section{State of the art}
%Contact planning is a key aspect of the more general motion planning problem in robotics~\cite{piano79}. Simultaneously solving a geometry problem (collision avoidance, joint limits), with a dynamics one (contact creation, and contact forces enabled) make the motion planning a non linear combinatorics problem of exponential complexity. An intuitive simplification consists in first computing the contacts locations, and then computing the motion that uses the contact locations without challenging them. 

Contact planning is a key aspect of motion planning in robotics~\cite{piano79}, as it requires simultaneously satisfying geometric constraints (collision avoidance, joint limits) and dynamic constraints (contact creation and contact forces). This combination makes motion planning a nonlinear combinatorial problem of exponential complexity. A common simplification in the literature is to first determine feasible contact locations and then plan the motion along these contacts. While intuitive, this separation limits the solution space and can compromise both optimality and robustness~\cite{tonneau2018efficient}.

Graph-based approaches address the combinatorial aspect by representing the problem as a search over contact states. Sampling-based methods~\cite{Bretl2006MotionPO,hauser2008motion,escande2013planning} explore feasible contact sequences probabilistically, offering flexibility and in some cases asymptotic guarantees of optimality. Early works solved a whole-body optimisation problem to generate motions and validate feasibility for each contact transition, resulting in frameworks too slow for interactive use. 
Other RRT-based methods for humanoid footstep planning generate steps by exploring discrete or projected feasible foot configurations, with swing-foot feasibility and collision checks (often using precomputed swept volumes)~\cite{LiuSZ12, perrin,cipriano}. These methods are probabilistically complete, but planning through narrow passages can require many rejected samples, potentially increasing computation time, particularly as the environment complexity grows.

Motion primitives of valid motions~\cite{Hauserprimitive2008} can be used to derive new motions through sampling or optimisation.  The idea of drawing motions from a database—essentially a probability distribution—appears in graphics~\cite{Choi03,pettre} and in reinforcement learning~\cite{deeploco}, although these methods are limited by the available data. Recent works aim to learn the feasible space through sampling~\cite{akizhanov2024learningfeasibletransitionsefficient,kumagailearning}.

Other approaches drastically reduce the contact search space by explicitly enumerating a finite set of contact transitions, defining a reduced action space efficiently explored with variations of the \astr algorithm~\cite{chestnuttkuffner,griffin2019footstep}. Since transitions are pre-defined, feasibility is implicitly guaranteed.  While effective in practice, this strategy restricts the search to a predefined discrete set of actions, trading coverage of the continuous reachability space for computational efficiency.

Optimization-based approaches typically use approximations of reachability constraints in the form of polytopes, which are easy to handle with off-the-shelf solvers~\cite{wieber,Winkler,orsolino}. Such approximations (and other more accurate~\cite{perrin}) have also been integrated into sampling-based methods to accelerate contact transition computations~\cite{tonneau2018efficient,deepgait}. Kumagai et al.~\cite{kumagai19} propose an \astr algorithm relying on polytopes, but they are eventually discretised, and do not handle reachability propagation beyond one step.

Finally, MIP methods combine continuous optimisation with guarantees of global optimality~\cite{deits2014footstep,aceituno2017mixed} while explicitly handling contact combinatorics. Despite the development of interactive solvers, MIP techniques remain computationally expensive and difficult to deploy in real time, as their runtime is hard to bound. These limitations motivate the development of methods that combine the efficiency and determinism of \astrns-based search with a continuous representation of reachability constraints, enabling fast and feasible contact planning. We propose to use recursive reachability~\cite{nas} to formulate continuous reachability constraints compatible with \astrns.

\section{Overview}

Given the current contact state of a biped robot and a target location, expressed either as a 3D position or a convex polyhedron (or polytope), \algo{} computes a sequence of feasible contact positions that allows the robot to reach the target in a minimum number of steps.

This is achieved in two stages:
\begin{itemize}
    \item Use a novel continuous \astr to identify a sequence of contact surfaces to walk upon (\refsec{sec:astar});
    \item Solve a Quadratic Program (QP), guaranteed to be feasible, to compute a sequence of contact positions that achieve the task (\refsec{sec:qp}).
\end{itemize}

%\subsection{\astr search}
\setlength{\tabcolsep}{2pt} % espace entre colonnes
\renewcommand{\arraystretch}{0} % pas d'espace vertical supplémentaire

% Réduit l’espace entre l’image et la sous-légende
\captionsetup[subfigure]{justification=centering, skip=0pt}

\begin{figure}[hb!]
\centering

% Ligne 1
\begin{subfigure}{0.48\linewidth}
    \centering
    \includegraphics[width=\linewidth]{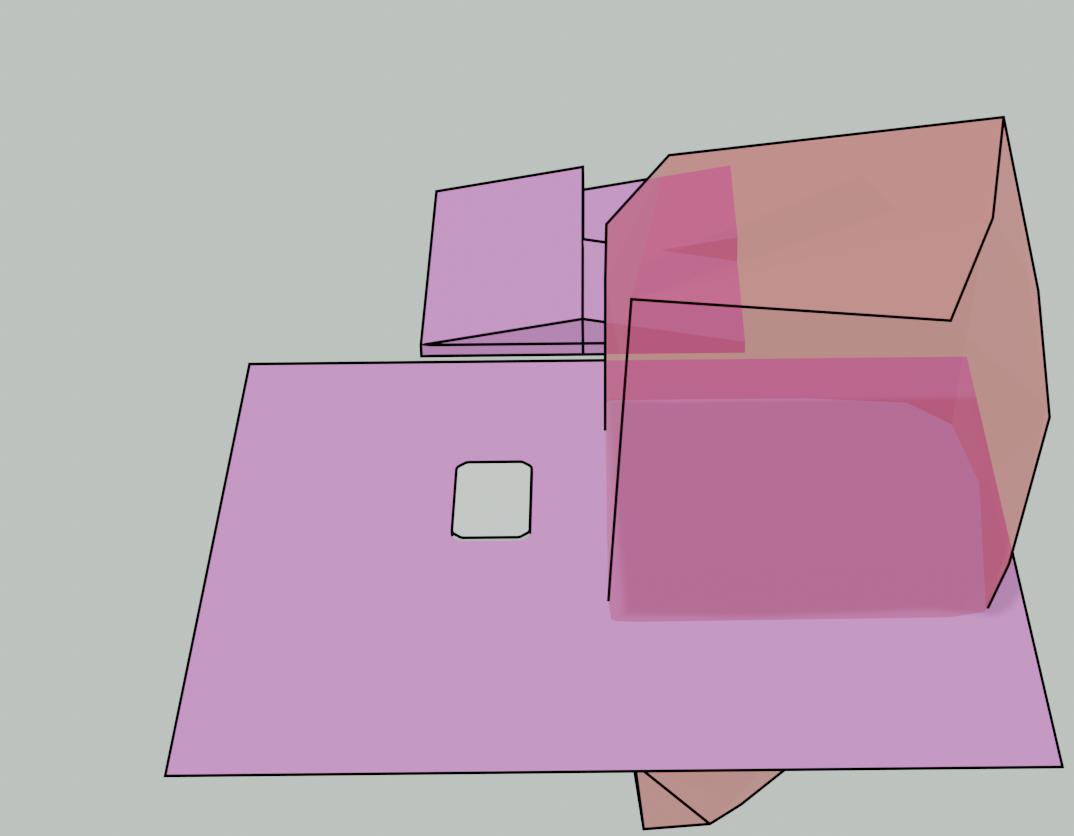}
    \caption{}
\end{subfigure}\hfill
\begin{subfigure}{0.48\linewidth}
    \centering
    \includegraphics[width=\linewidth]{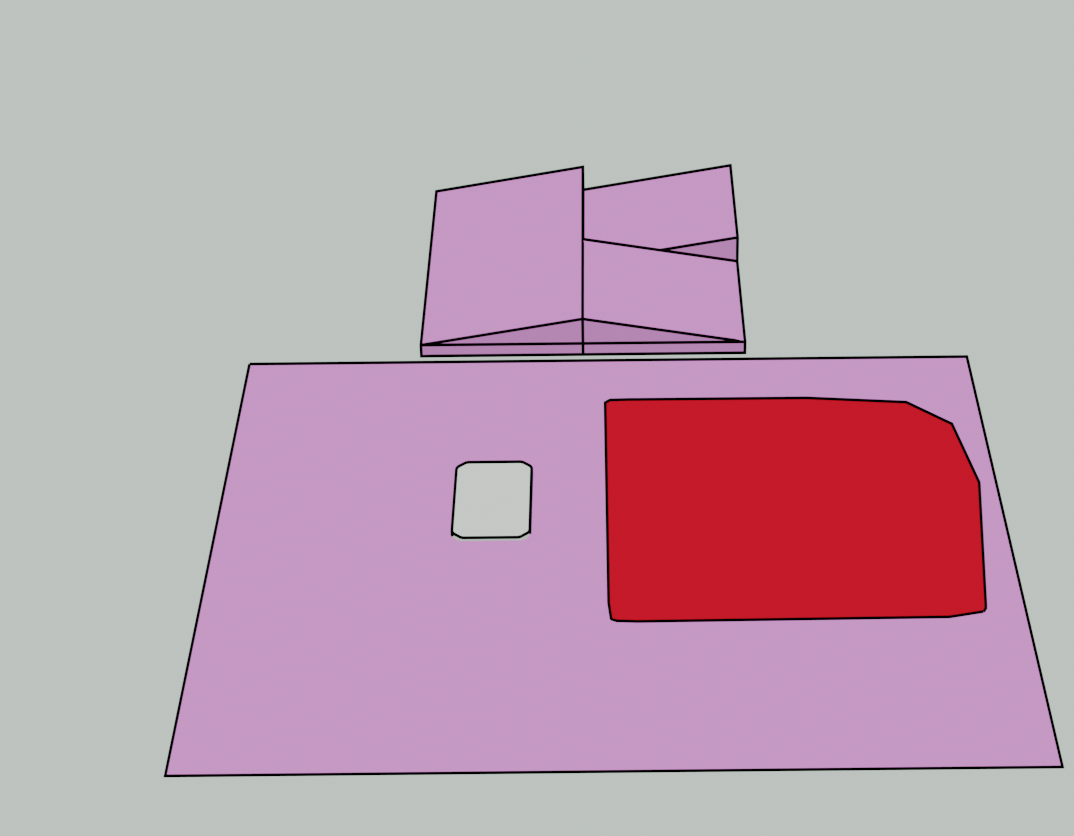}
    \caption{}
\end{subfigure}

% Ligne 2
\begin{subfigure}{0.48\linewidth}
    \centering
    \includegraphics[width=\linewidth]{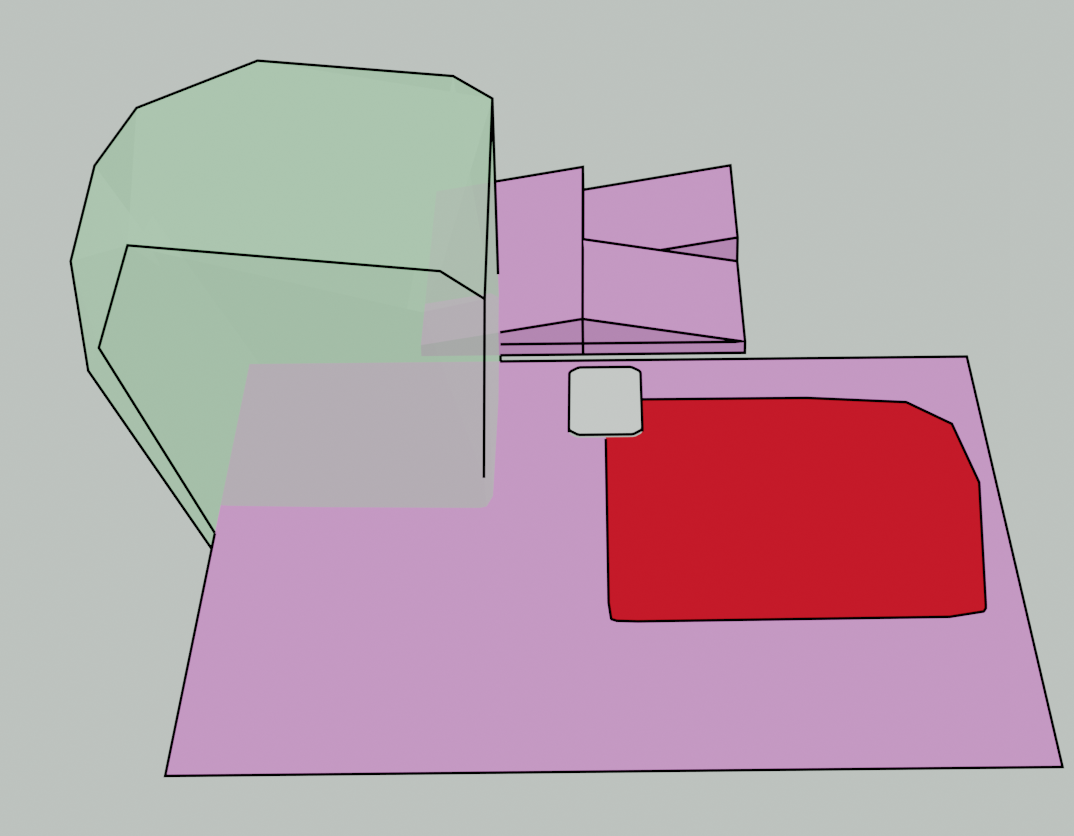}
    \caption{}
\end{subfigure}\hfill
\begin{subfigure}{0.48\linewidth}
    \centering
    \includegraphics[width=\linewidth]{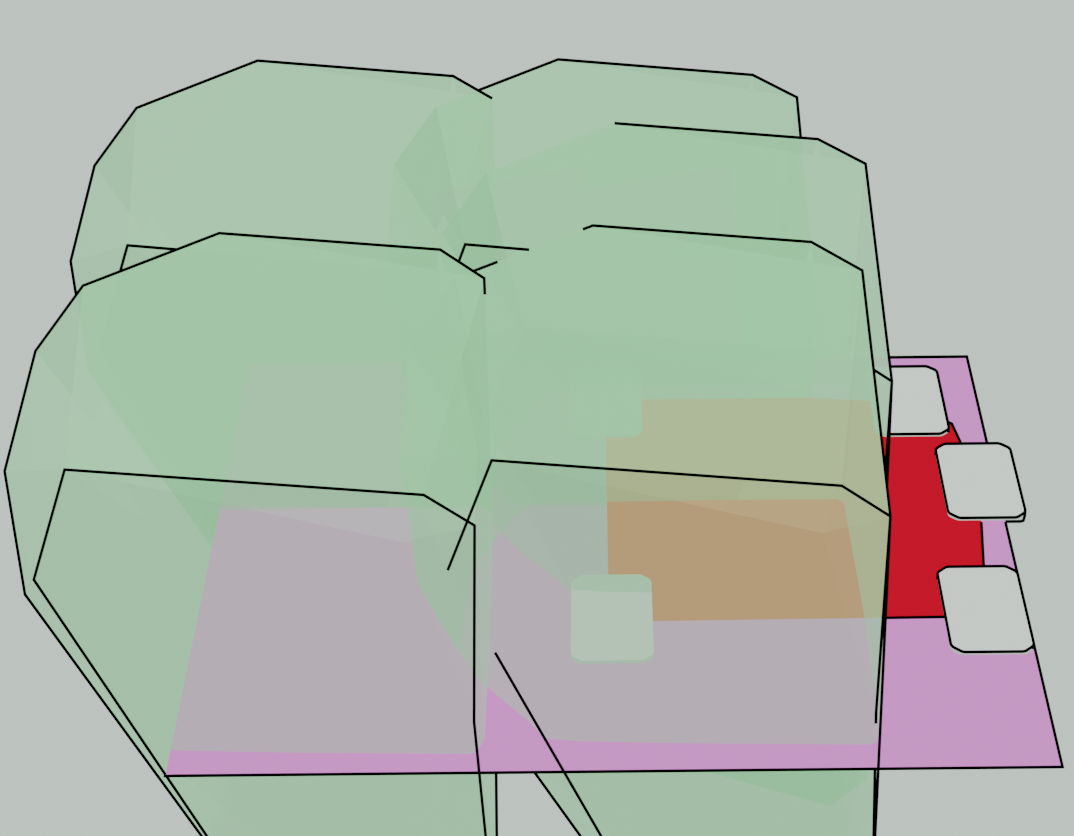}
    \caption{}
\end{subfigure}

% Ligne 3
\begin{subfigure}{0.48\linewidth}
    \centering
    \includegraphics[width=\linewidth]{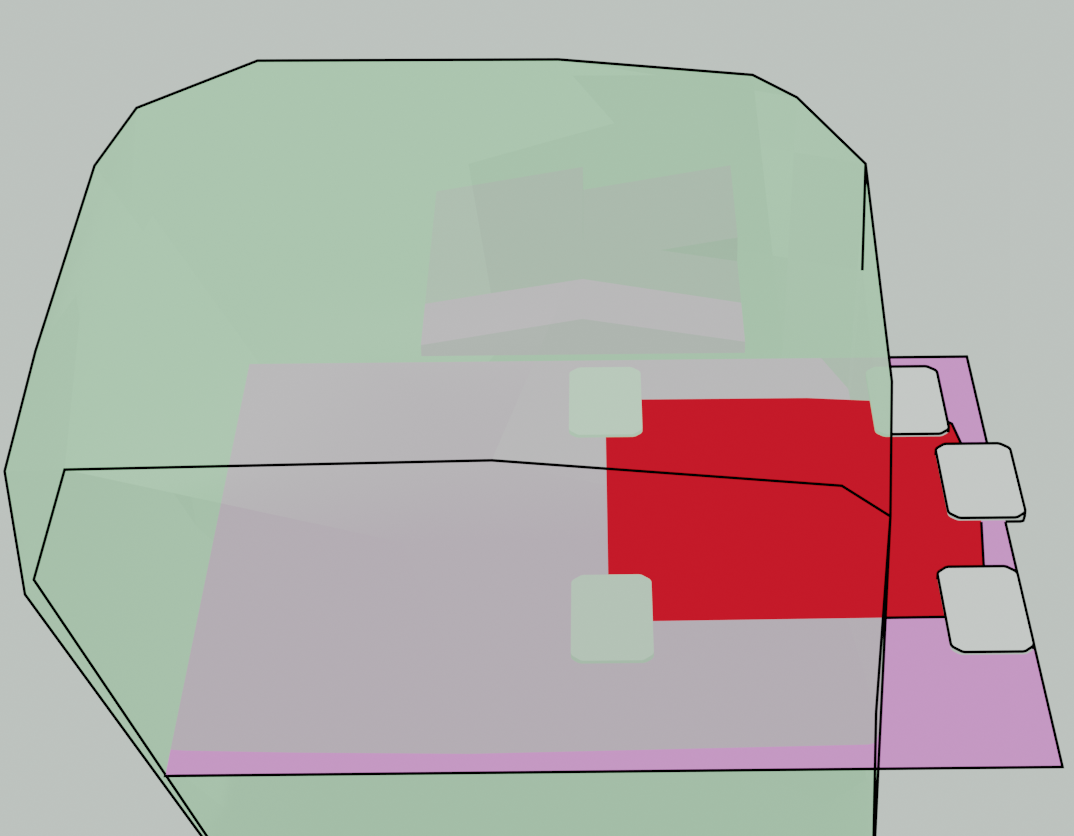}
    \caption{}
\end{subfigure}\hfill
\begin{subfigure}{0.48\linewidth}
    \centering
    \includegraphics[width=\linewidth]{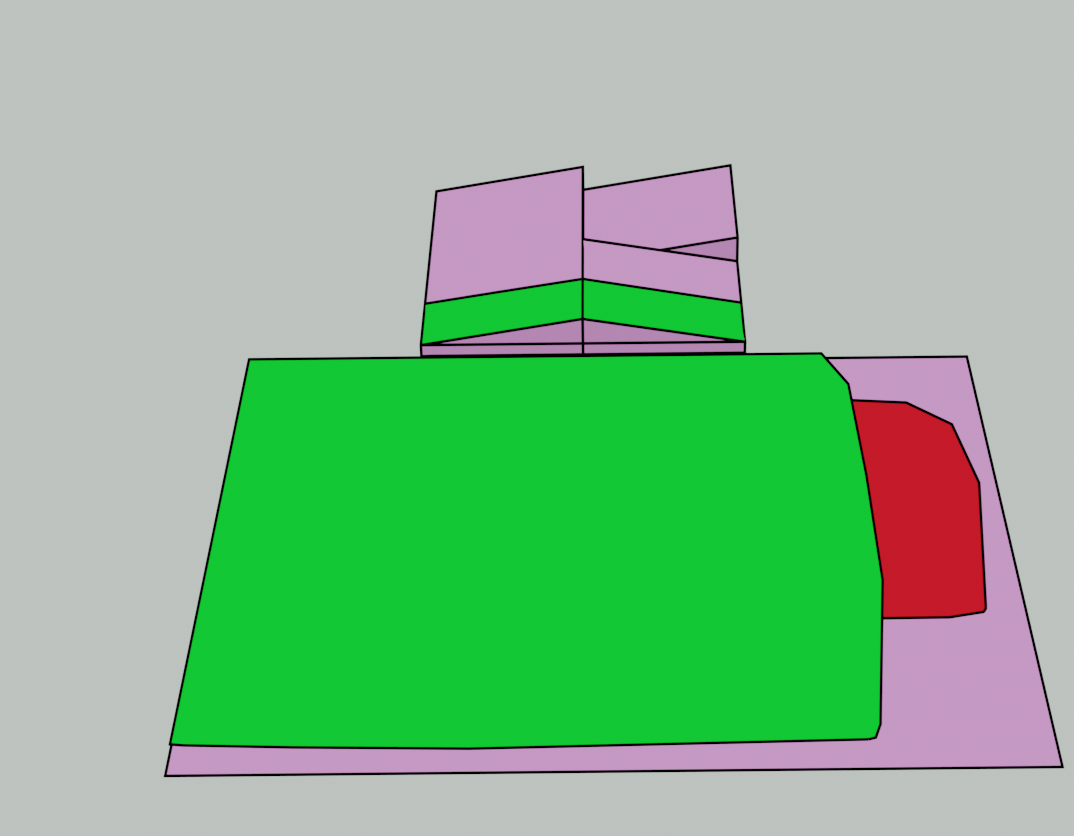}
    \caption{}
\end{subfigure}

\caption{The pictures are taken from the back of the robot (the forward direction is towards the stepping stones). Given the position of a foot (grey rectangle) and assuming no rotation,  we can compute the reachability constraints for the other foot as a convex polytope (red for constraints on the right foot, green for the left foot). The set of positions reachable from the red polygon (b) is obtained by computing the Minkowski sum of all reachability polytopes for all positions in the polygon (c, d, e). This is computed efficiently as the convex hull of the reachability polytopes of the extreme points (d). Not all extreme points are shown in (d) for readability. The union of the 3 green polygons in (f) is the 2 step-reachable set and corresponds to 3 nodes added to the \astr graph.}
\label{fig:expand}
\end{figure}

The node expansion step and the heuristic of the \astr are implemented as follows. A node in the graph is associated with a convex contact surface (\reffig{fig:expand}-b - red) that describes all potential positions for a given effector that can provably be reached from the parent node in 1 step.

During the node expansion phase 
(\refsec{sec:expansion}), the set of reachable positions is computed as a convex polyhedron (polytope in the remainder of this paper - \reffig{fig:expand}-a - red), intersected with identified convex contact surface candidates (\reffig{fig:expand}-b). For each surface that the polytope intersects, one node is created and added to an active set of nodes (\reffig{fig:expand}-f). The possible yaw orientations of the base are discretised~\cite{deits2014footstep, griffin2019footstep} and thus contribute to the branching factor.

\astr selects a node to expand based on the number of steps required to reach the node from the start, along with a cost-to-go heuristic that computes the minimal Euclidean distance between a position in the node and the target using the EPA algorithm~\cite{vandenBergen2001}, implemented in coal~\cite{coalweb}. No other components of \astr are modified. The node expansion uses the n-step reachability formulation proposed in~\cite{nas}, which we extend by allowing to include rotation in the planning.

%\subsection{Computation of the contact sequence with a QP \stn{section}}
%Once a sequence of contact surfaces has been identified, a simple QP can be written to constrain each step location to belong to the surface assigned to its specific phase. A postural quadratic cost and a `safety` cost to avoid stepping on the edges of the surfaces if possible is added.  \jwn{we do not have a posture cost, but we ask the footstep to be close to the patch center (not very useful i think); the safety margin is included in a star search.}

\section{Assumptions and definitions}
\label{sec:assumptions}

\newcommand{\pol}[1]{\mathcal{#1}}
\newcommand{\pole}[2]{\mathcal{#1}^{#2}} 
\newcommand{\rkl}[0]{^{r}\mathcal{K}_{l}} 	
\newcommand{\lkr}[0]{^{l}\mathcal{K}_{r}} 	
\newcommand{\ral}[0]{^{r}\mathcal{A}_{l}} 	
\newcommand{\lar}[0]{^{l}\mathcal{A}_{r}} 
\newcommand{\env}[0]{\mathcal{S}} 	
\newcommand{\envj}[1]{\mathcal{S}^{#1}} 	
\newcommand{\tg}[0]{\mathcal{G}} 	

\newcommand{\rotm}{\vc{Q}} 
\newcommand{\fe}[1]{\mathcal{F}_{#1}} 
\newcommand{\re}[1]{\mathcal{R}_{#1}} 
\newcommand{\kee}[0]{^{e}{\mathcal{A}}_{\overline{e}}}
\newcommand{\keem}[0]{^{e}{\mathbf{K}}_{\overline{e}}}
\newcommand{\ree}[0]{^{\vc{p}_e}\mathcal{R}}
\newcommand{\reem}[0]{^{\vc{p}_e}\mathbf{R}}
\newcommand{\pne}[0]{\vc{p}_{\overline{e}}}
\newcommand{\pe}[0]{\vc{p}_{{e}}}
\newcommand{\fer}[0]{^{\pol{R}_e}\mathcal{F}}

\newcommand{\tre}[0]{\mathcal{T}} 

\newcommand{\rer}[0]{^{\pol{R}_e}\mathcal{R}}
\newcommand{\reb}[0]{\pol{R}_e}
\newcommand{\rem}[0]{\vc{R}_e}
\newcommand{\rerm}[0]{^{\pol{R}_e}\mathhbb{R}}

Our framework allows for contact plans that involve stepping on the same surface several times with the same foot. However, we assume that once a foot has created a contact on a new surface, it will not step again on the previous surface. Rare cases where this could be a problem could be addressed by further decomposition of very long surfaces. Our results are currently limited to bipedal locomotion.

\subsection{Polytope-based representation of constraints}
The environment and other constraints are described in the form of convex 3D polytopes. A polytope  $\mathcal{P}$ is  defined as
\begin{equation}\label{eq:pol}
\begin{aligned}
    \mathcal{P} := \{\mathbf{p} \in \mathbb{R}^3 | \exists \bm{\lambda} \in \mathbb{R}^{+d},   \|\bm{\lambda}\|_1 = 1 \wedge \mathbf{p} = \mathbf{P} \bm{\lambda} \}  \,.
\end{aligned}
\end{equation} where $\vc{P}$ is a matrix of dimension $3 \times d $ obtained by stacking the $d$ extreme points of the polytope. Any point $\vc{p}$ that can be obtained as a convex combination of the extreme points of $\mathcal{P}$ belongs to the polytope.

\textit{The environment} is defined as the union of  $m$ disjoint convex contact surfaces $\env = \bigcup_{j=1}^{m} \envj{j}$, as commonly done~\cite{deits2014footstep,Grandia-perceptive-through-NMPC}. Each $\envj{j}$ is a polygon in 3D (\reffig{fig:expand}-purple top surfaces). No assumptions are made on the orientation of the surfaces.

\subsection{Reminder on reachability computation}
\label{sec: expansion}
We now provide a reminder on the notion of n-step reachability but refer the reader to ~\cite{nas} for additional details.

\begin{figure}[ht]
\centering
\includegraphics[width=0.35\linewidth]{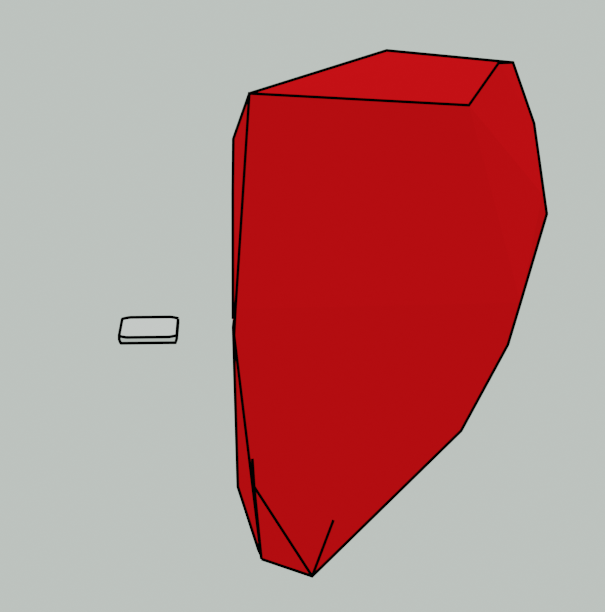}
\caption{Reachable set (red polytope) for the right foot with respect to the left foot (rectangle), seen from the back of the robot. To avoid leg crossing the right foot is constrained to be ``on the right'' of the left foot by construction of the set.}
\label{fig:kin}
\end{figure}

\paragraph{1-step reachability from a 3D foot position}{The kinematic constraints} of the robot are linearised~\cite{wieber,Winkler}. We define as $ \rkl$ the polytope describing all reachable positions for the left foot in the frame of the right foot. $\lkr$ describes the constraints for the right foot in the frame of the left foot (\reffig{fig:kin}). By definition, $ \rkl$ and $ \rkl$ are the 1-step reachable sets for a given position of the right foot (respectively the left foot), expressed in the local foot frame $^{e}\vc{T}$. The location of the  1-step reachable polytope $\mathcal{R}_e$ is simply obtained through the transformation of the extreme points $^{e}\vc{K}_{\overline{e}}$ by  ${^{e}\vc{T}}^{-1} $.

\paragraph{1-step reachability from a polytope}
If $\mathcal{R}_e$ is a polytope that contains all the possible positions for an effector $e$ on the same contact surface (thus assuming the same contact normal), we are interested in computing all the positions that can be reached for the other effector $\overline{e}$ from  $\mathcal{R}_e$ (\reffig{fig:expand}-c, d, e). If the foot yaw orientation is constant, this set is also a convex polytope $\rer$, given by the Minkowski sum~\cite{nas}: 
\begin{equation}\label{eq:rer}
\begin{aligned}
    \rer:= \{  \rotm {^{e}\vc{K}_{\overline{e}}} \ \bm{\lambda} +    \rem\bm{\lambda}_1, \ ||\bm{\lambda}|| = ||\bm{\lambda}_1|| = 1 \}  \,.
\end{aligned}
\end{equation}
with $\bm{\lambda}$ and $\bm{\lambda}_1$ positive valued-vectors of appropriate dimension, and $\rotm$  the rotation matrix that matches the current yaw and contact surface rotation.

\paragraph{Rotations introduce a combinatorics}
If the foot yaw angle is variable, the reachable sets are no longer convex. To remove the assumption that the yaw is constant, we discretise the possible orientations, as done in the state of the art~\cite{deits2014footstep, griffin2019footstep}. This results in a conservative approximation of the search space and introduces a combinatorics which, as for the contact surface choice, is handled by the \astr.

\section{\astr implementation}
\label{sec:astar}
\refalg{alg:astar_standard} provides the pseudocode for \astr. The pseudocode uses the same syntax proposed by~\cite{griffin2019footstep}. The specific details of our approach are highlighted in \textcolor{blue}{blue}.

\begin{algorithm}
\caption{A* Search Algorithm}
\label{alg:astar_standard}
\begin{algorithmic}[1]
\STATE addStartNodeToQueue()
\WHILE{hasNodesToCheck()}
    \STATE nodeToExpand = getCheapestNode();
    \IF{\textcolor{blue}{nodeAlreadyExpanded() // 
\textit{\refsecs{sec:nodeAlreadyExpanded}}}} 
        \STATE  skipToNextNodeInQueue();
    \ENDIF
    \IF{hasReachedTheGoal() or hasTimedOut()}
        \STATE stopSearch();
    \ENDIF
    \STATE \textcolor{blue}{childNodesToCheck = expandNode(); // \textit{\refsecs{sec:expansion}}}
    \FOR{childNode $\in$ childNodesToCheck}     
        \IF{\textcolor{blue}{hasContactSurfaceNotBeenLeft() // 
\textit{\refsecs{sec:hasContactSurfaceNotBeenLeft}}}} 
            \STATE continue // skip node
        \ENDIF        
            \textcolor{blue}{\STATE costOfEdge = getCostOfEdge();    // \textit{\refsecs{sec:cost}}}
            \STATE addEdgeToGraph(childNode, costOfEdge);
            \textcolor{blue}{\STATE costToGoal = estimateCostToGoal(); // \textit{\refsecs{sec:cost}}}
            \STATE nodeCost = costOfPath(childNode) + costToGoal;
            \STATE addNodeToQueue(childNode, nodeCost);
    \ENDFOR
\ENDWHILE
\STATE footstepPlan = getBestPathToEndNode();
\end{algorithmic}
\end{algorithm}

\subsection{Reminder on the \astr algorithm}
\astr uses a priority queue that sorts nodes based on the cost to reach the node from the initial node and an estimation of the remaining cost to the target (line 18). The nodes are connected in a graph (line 16).
At each iteration, the most promising node is selected (line 3). If this node is terminal (line 7), the solution node sequence is obtained by iterating through the parents of the graph starting from the current node (line 22). Otherwise, the node is expanded (line 10), and the cost of the new nodes is evaluated before they are added to the priority queue (lines 15-19).

\subsection{Specific implementation details}
\subsubsection{Node structure}
We use the node structure proposed in~\cite{nas}, shown in \rstruct{alg:strcut}, modified to include rotation information. 
A node is composed of an identifier for the end-effector currently in contact, the contact surface, and the subset of the surface covered by the node (\reffig{fig:expand} - b). It also contains a link to its parent in the graph and the yaw angle associated with the contact foot. A node has only one parent.

\setcounter{algorithm}{0}

\captionsetup[algorithm]{labelformat=empty}
\begin{algorithm}
\caption{\textbf{struct 1} \textsc{Node}}
\begin{algorithmic}%[1] % uncomment for line numbers
%\Struct{Node}
%\State \textbf{struct} \textsc{Node}
  \STATE \hspace{1em} $effectorId$: \textsc{enum} 
  \STATE \hspace{1em} $parent$: \textsc{Node*} // parent node
  \STATE \hspace{1em} $surfaceId$: \textsc{int}
  \STATE \hspace{1em} $extremePoints$: \textsc{Point list} //polygon description
 % \EndStruct 
  \STATE \hspace{1em} $yaw$ : \textsc{double} // contact yaw angle in radians
\end{algorithmic}
 \label{alg:strcut}
\end{algorithm}

\subsubsection{expandNode()}
\label{sec:expansion}
The $extremePoints$ of the current node define the polytope $\mathcal{R}_e$ of potential positions associated with the node. \refalg{alg:expand} provides the pseudocode of the expansion step. For each discretised yaw angle, we compute $\rer$ using \refeq{eq:rer} (lines 3 and 4) and intersect it with the surfaces of the environment (line 6). For each non-empty intersection, we generate a new node for which $effectorId$ is the other foot, $parent$ is the current node, and $extremePoints$ are the obtained intersection (line 8).

\begin{algorithm}
\caption{expandNode function}
\label{alg:expand}
\begin{algorithmic}[1]
     \STATE nodeList $= []$;
    \FORALL{$\theta_i$ in discretisedYaws}  
        \STATE currentYaw $= yaw + \theta_i $;
        \STATE $\re{e_i} =$ rotateKinConstraint($surfaceId$, currentYaw)
        \STATE $\rer{_i} = $Minkowski$(\re{e_i}, effectorId)$
        \FORALL{surface $\mathcal{S}_j$ in $\pol{S}$}
            \STATE reachableContact = intersect($\rer{_i}$, $\mathcal{S}_j$)        
            \IF{notEmpty(contact)}
                \STATE  nodeList.add(createChildNode($effectorId$, this, j, reachableContact, currentYaw));
            \ENDIF
        \ENDFOR
    \ENDFOR
    \RETURN return nodeList
\end{algorithmic}
\end{algorithm}

\subsubsection{$\textrm{nodeAlreadyExpanded()}$}
\label{sec:nodeAlreadyExpanded}
This function checks whether the current node corresponds to any previously expanded node. It returns $True$ if the $effectorId$ and $surfaceID$ are identical, and if the Euclidean distance between both the centre and the perimeter of the patches is below a user-defined threshold, set empirically to 2 $cm$.

% This function returns $True$ if the $effectorId$, $surfaceID$ of the current node and one of its parents match, \textbf{and} \jwn{if the center and perimeters of the patches are similar, e.g., 0.02m.}
% if the $extremePoint$ of both nodes match to a given tolerance.  \stn{@jiayi how much}

\subsubsection{$\textrm{hasContactSurfaceNotBeenLeft()}$}
\label{sec:hasContactSurfaceNotBeenLeft}
As explained in \refsecs{sec:assumptions}, we prevent revisiting the same surface with the same foot after having left it. This step empirically prevents long expansions that often prove suboptimal.

\subsubsection{Cost computation}
\label{sec:cost}
The cost of a node is defined as the number of steps required to reach the node from the initial position. Thus, $\textrm{getCostOfEdge()}$ always returns 1. We do not impose a penalty on the rotation to illustrate the interest of our approach, as discussed in 
\refsec{sec:experiments}. In the companion video, for qualitative purpose we also demonstrate videos that include a small cost to penalise yaw rotation.

The cost-to-go estimateCostToGoal() is an estimation of the cost to reach the target given the current node. A typical cost for a discretised \astr is the Euclidean distance between the node position and the target. 
We propose a more generic formulation, more suited to our problem, since either the target or the node can be described as a polytope.
We use the EPA algorithm \cite{vandenBergen2001}, an extension of the GJK algorithm that computes efficiently the minimum distance between 2 polytopes (or a polytope and a point). 
This heuristic can be considered as a scaling of the lower bound on minimum number of steps to the target based on the maximum stride length. This scaling makes the heuristic not admissible and classifies \algo{} as a weighted \astrns. We sacrifice the guarantee of global optimality in favour of a more efficient search because our experiments show that the optimum is always empirically found by \algo{}. 

\section{Footstep computation}
\label{sec:qp}
\astr outputs is a list of contact surfaces $\mathcal{F} = [\pol{F}_1, \dots, \pol{F}_l ]$ on which the robot can step to reach the goal, along with the corresponding rotation matrices $\mathbf{A} = [\vc{A}_1, \dots, \vc{A}_l ]$  with $l$ the total number of steps. The foot orientation is thus fixed by the \astr.  The footstep sequence $\vc{X}$ is obtained through the linearly constrained program:

\begin{equation}\label{eq:qp}
\begin{aligned}
    \textbf{find} \quad & \mathbf{X} = [\mathbf{x}_1,\cdots,\mathbf{x}_{l}] \in \mathbb{R}^{3\times l} \\ 
    \textbf{min} \quad & c(\mathbf{X}) \\
    \textbf{s.t.}
    \quad & \forall i, 1 \leq i \leq l :& \\ 
    \quad & \quad \quad \mathbf{x}_{i} \in  ^{\vc{x}_{i-1}}\mathcal{R}  \cap \pol{F}_{i} 
\end{aligned}
\end{equation}

where $\vc{x}_0$ is a starting foot position, $c$ a cost function and 

\begin{equation}\label{eq:pol2}
\begin{aligned}
    ^{\vc{x}_{i-1}}\mathcal{R} := \{\mathbf{p} \in \mathbb{R}^3 | \exists \bm{\lambda} \in \mathbb{R}^{+d},  \\  \|\bm{\lambda}\|_1 = 1 \wedge \mathbf{p} = \vc{A}_{i-1} (^{e(i-1)}\vc{K}_{e(i)}) \bm{\lambda} + \vc{x}_{i-1}\}  \, ,
\end{aligned}
\end{equation}
with $e(i)$ the appropriate effector for the phase. In our implementation, we use the inequality forms of both $^{\vc{x}_{i-1}}\mathcal{R}$ and $\pol{F}_{i}$ for efficiency~\cite{fukudadouble}.

In our experiments, $c$ minimises the average stride length between each step, i.e.  $c(\vc{X}) = \sum_{i=2}^{l}(x_i-x_{i-2})^2$. 
\subsection*{Foot placement robustness}
Our algorithm allows stepping on the edge of contact surfaces, where the entire foot is not guaranteed to fit. Although this feature contributes to the completeness of the approach under our assumptions~\cite{griffin2019footstep}, robustness can be obtained by avoiding these positions if possible.
We use the inequality representation \cite{fukudadouble} of the  $\pol{F}_{i}$:

\begin{equation}\label{eq:ineq}
\begin{aligned}
    \mathcal{F}_i := \{\mathbf{p} \in \mathbb{R}^3 | \vc{B}_i \mathbf{p}  \leq  \vc{b}_i \} 
\end{aligned}
\end{equation}

where $\vc{B}_i$ are matrices and vectors of appropriate dimensions $\vc{b}_i$, ``normalised'' such that the norm of each line of $\vc{B}_i$ is equal to 1. We then introduce a single slack variable $\alpha \in \mathbb{R}^+$ and solve the following program:  

\begin{equation}\label{eq:qp}
\begin{aligned}
    \textbf{find} \quad & \alpha, \mathbf{X} = [\mathbf{x}_1,\cdots,\mathbf{x}_{l}] \in \mathbb{R}^{3\times l} \\ 
    \textbf{min} \quad & c(\mathbf{X}) - 10 \alpha \\
    \textbf{s.t.}
    \quad & \forall i, 1 \leq i \leq l :& \\ 
    \quad & \quad \quad \mathbf{x}_{i} \in  ^{\vc{x}_{i-1}}\mathcal{R} \\ 
    \quad & \quad \quad \vc{B}_i \mathbf{x}_i + \vc{1}_i\alpha \leq  \vc{b}_i 
\end{aligned} 
\end{equation}
where $\vc{1}_i$ is a vector of ones of appropriate dimension.
If $c = 0$, this linear program returns a contact sequence such that the lowest distance to a surface edge is maximised. We empirically scale $\alpha$ by a factor 10 in our experiments. 

\section{Experiments}

\label{sec:experiments}

\begin{table*}[!hb]
\centering
\renewcommand{\arraystretch}{1.4}
\setlength{\tabcolsep}{4pt}
\resizebox{\textwidth}{!}{%
\begin{tabular}{|l|l|c|c|c|c|c|c|c|c|c|c|c|c|c|c|c|}
\hline
\multicolumn{2}{|c|}{\textbf{Method}} & 
\multicolumn{5}{c|}{\textbf{Stairs}} & 
\multicolumn{5}{c|}{\textbf{Local Minima}} & 
\multicolumn{5}{c|}{\textbf{Narrow Passage}} \\
\hline
 & & \multicolumn{3}{c|}{Computation time (ms)} & \# of Nodes & \# of Steps 
   & \multicolumn{3}{c|}{Computation time (ms)} & \# of Nodes & \# of Steps 
   & \multicolumn{3}{c|}{Computation time (ms)} & \# of Nodes & \# of Steps \\
\cline{3-5}\cline{8-10}\cline{13-15}
  & & A* & QP & Total & & 
    & A* & QP & Total & & 
    & A* & QP & Total & & \\
\hline
\multirow{2}{*}{\algo{} (ours)} 
 & without rotation 
 & 4.67$\pm$0.17 & 8.51$\pm$0.88 & 13.18$\pm$0.91 & 16 & \color{ForestGreen}{15}  %stair
 & {\color{ForestGreen}8.08$\pm$0.26} & 33.18$\pm$3.03 & {\color{ForestGreen}41.27$\pm$2.85} & 42 &{\color{ForestGreen}25} %local minima
 & Fail & - & - & - & - \\ %narrow
\cline{2-17}
 & with rotation 
 &{\color{ForestGreen}10.00$\pm$0.15} & 4.74$\pm$0.60 & {\color{ForestGreen}14.75$\pm$0.53} & 33 & \color{ForestGreen}{11} %stair
 & {\color{ForestGreen}33.66$\pm$0.9} & 14.61$\pm$0.92 & {\color{ForestGreen}48.27$\pm$0.70} & 92 & {\color{ForestGreen}19} %local minima
 & {\color{ForestGreen}60.41$\pm$0.46} & 64.72$\pm$1.22 & {\color{ForestGreen}125.13$\pm$1.43} & 88 & {\color{ForestGreen}29} \\ %narrow
\hline
 % & with rotation (yaw cost)   
 % &{\color{ForestGreen}10.00$\pm$0.15} & 4.74$\pm$0.60 & {\color{ForestGreen}14.75$\pm$0.53} & 33 & \color{ForestGreen}{11} %stair
 % & {\color{ForestGreen}33.66$\pm$0.9} & 14.61$\pm$0.92 & {\color{ForestGreen}48.27$\pm$0.70} & 92 & {\color{ForestGreen}19} %local minima
 % & {\color{ForestGreen}60.41$\pm$0.46} & 64.72$\pm$1.22 & {\color{ForestGreen}125.13$\pm$1.43} & 88 & {\color{ForestGreen}29} \\ %narrow
% \hline
\multirow{2}{*}{Discretized A*} 
 & without rotation 
 & {\color{ForestGreen}4.32$\pm$0.07} & - & {\color{ForestGreen}4.32$\pm$0.07} & 102 & \color{ForestGreen}{15}
 & 61.81$\pm$0.29 & - & 61.81$\pm$0.29 & 728 & {\color{ForestGreen}25} 
 & Fail & - & - & - & - \\
\cline{2-17}
 & with rotation
 & 38.50$\pm$0.95 & - & 38.50$\pm$0.95 & 253 & \color{red}{15} 
 & 6626.53$\pm$45.61 & - & 6626.53$\pm$45.61 & 23233 & {\color{red}23} 
 & 1456.19$\pm$15.15 & - & 1456.19$\pm$15.15 & 1851 &  {\color{red}41}      \\
\hline
%  & with rotation (yaw cost)   
%  & 38.50$\pm$0.95 & - & 38.50$\pm$0.95 & 253 & \color{red}{15} 
%  & 6626.53$\pm$45.61 & - & 6626.53$\pm$45.61 & 23233 & {\color{red}23} 
%  & 1456.19$\pm$15.15 & - & 1456.19$\pm$15.15 & 1851 &  {\color{red}41}      \\
% \hline
MIP (optimal) & without rotation 
& - & - & 83.68$\pm$5.76  & 1 & 15 
& - & - & 228.44$\pm$8.29 & 1 & 25 
& Fail & - & - & - & - \\
\hline
MIP (30 steps) & without rotation 
& - & - & 400.55$\pm$8.83  & 294  & 30
& - & - & 916.88$\pm$17.53 & 1055 & 30
& Fail & - & - & - & - \\
\hline
\end{tabular}%
}
\caption{Computation times and number of nodes expanded on all scenarios different instances of planners}
\label{table:result}
\end{table*}

We compare \algo{} with a discretised \astr and a Mixed-Integer solver in 6 scenarios for the Talos robot~\cite{stasse2017talos}.

\subsection{Implementation details for the \astr solvers}
Both discretised \astr and \algo{} are implemented in C++ using the same code base, and differ only in the expansion step and node implementation. We discretise $^{e}\vc{K}_{\overline{e}}$ using a granularity of 0.05$m$ as in~\cite{griffin2019footstep}. We measure the total time spent in the \astr for each instance. Additionally for \algo{}, we measure the time to solve the QP in \refeq{eq:qp}. We describe the rotation using $10\degree$ increments from $-30\degree$ to $30\degree$.

\subsection{Implementation details for the MIP}
We use a public implementation of the MIP solver~\cite{song2020solving}, based on Gurobi~\cite{gurobi}. The solver does not allow for yaw rotation, so MIP was not tested in these scenarios.
Because MIP formulations require a fixed horizon (i.e. a fixed number of steps), we tested two formulations. In the first instance, the planning horizon is fixed to the known minimal number of steps obtained with a global solver~\cite{nas}. This is a favourable instance for the MIP, such that the pre-solve operation~\cite{gurobiprimer} eliminates all branches but one, effectively cancelling the combinatorics. In the second instance, which is a more realistic use of the solver, we set a fixed horizon of 30 steps. We only measure the time actually spent in the resolution of the MIP problem, and not its initialisation.

\subsection{Objective functions}
The MIP cost is set to 0 as we solve a feasibility problem to favour the MIP in the experiments. 
The discretised \astr uses the number of steps as the path cost and the Euclidean distance between the current step and the target as the cost-to-go estimation, which is equivalent to the cost for \algo{}.

\subsection{Scenarios}
\begin{figure*}[hbp]
    \centering
    % ligne 1
    \includegraphics[width=0.3\textwidth]{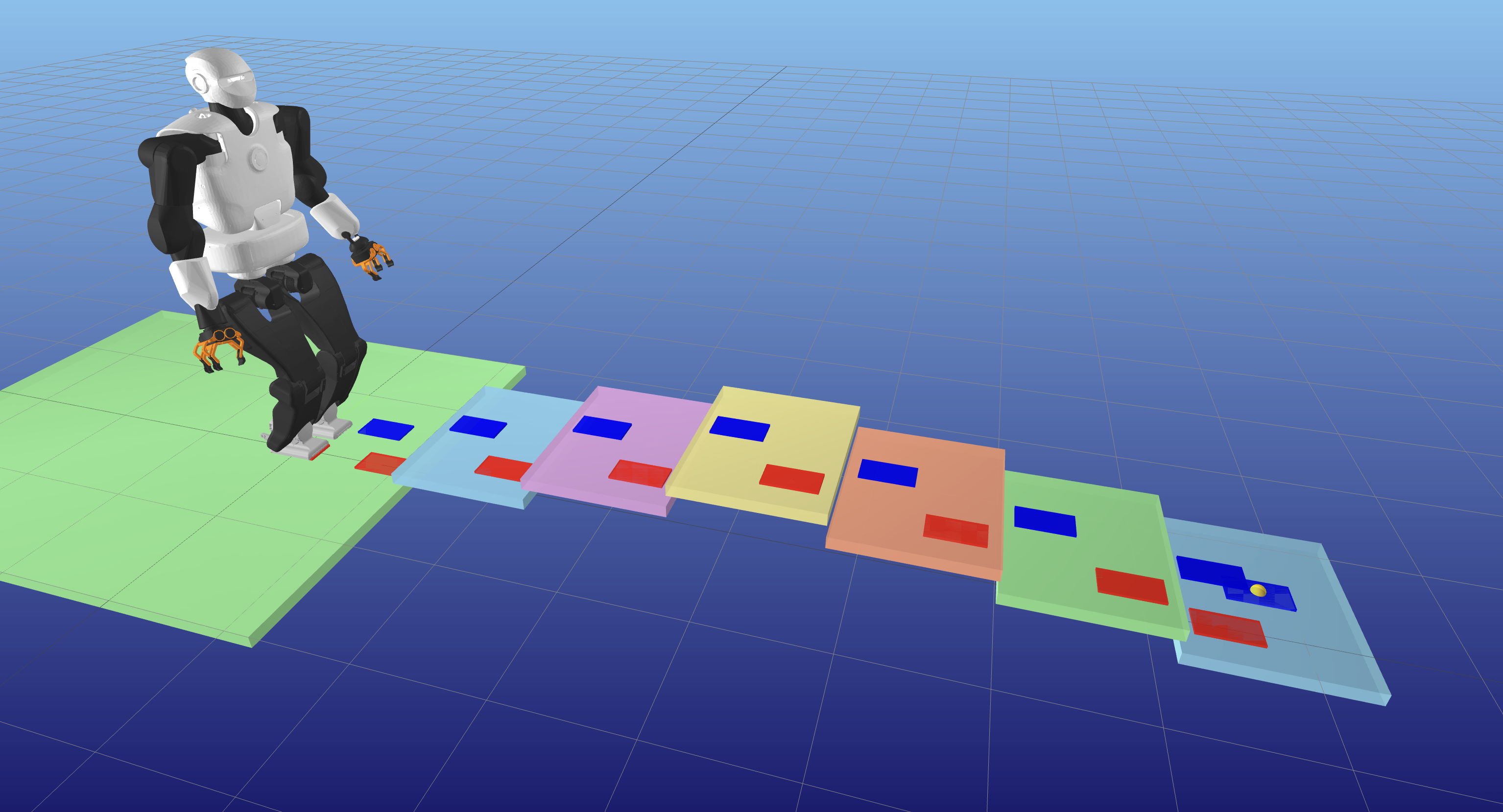}
    \includegraphics[width=0.3\textwidth]{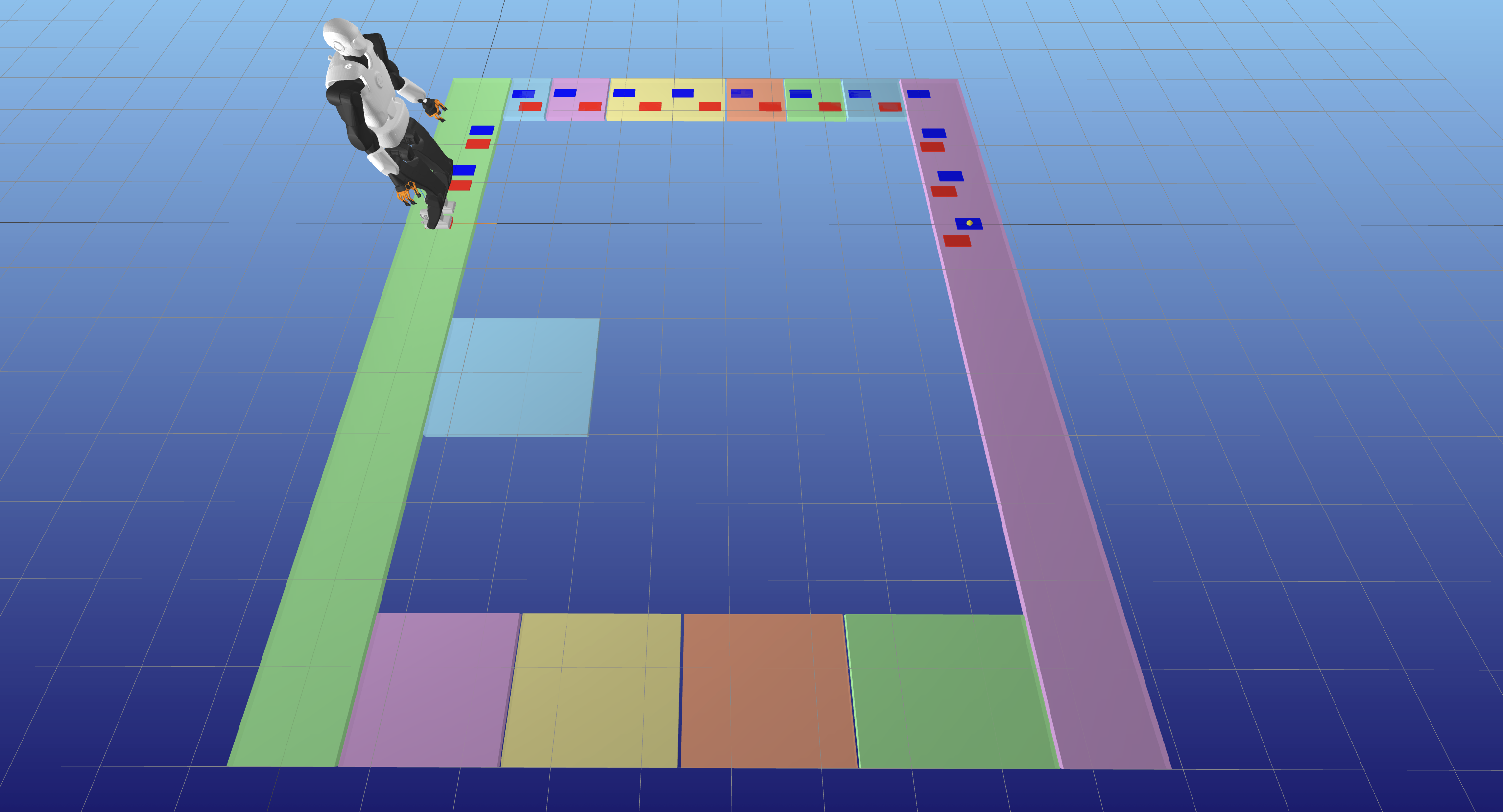}
    \phantom{\includegraphics[width=0.3\textwidth]{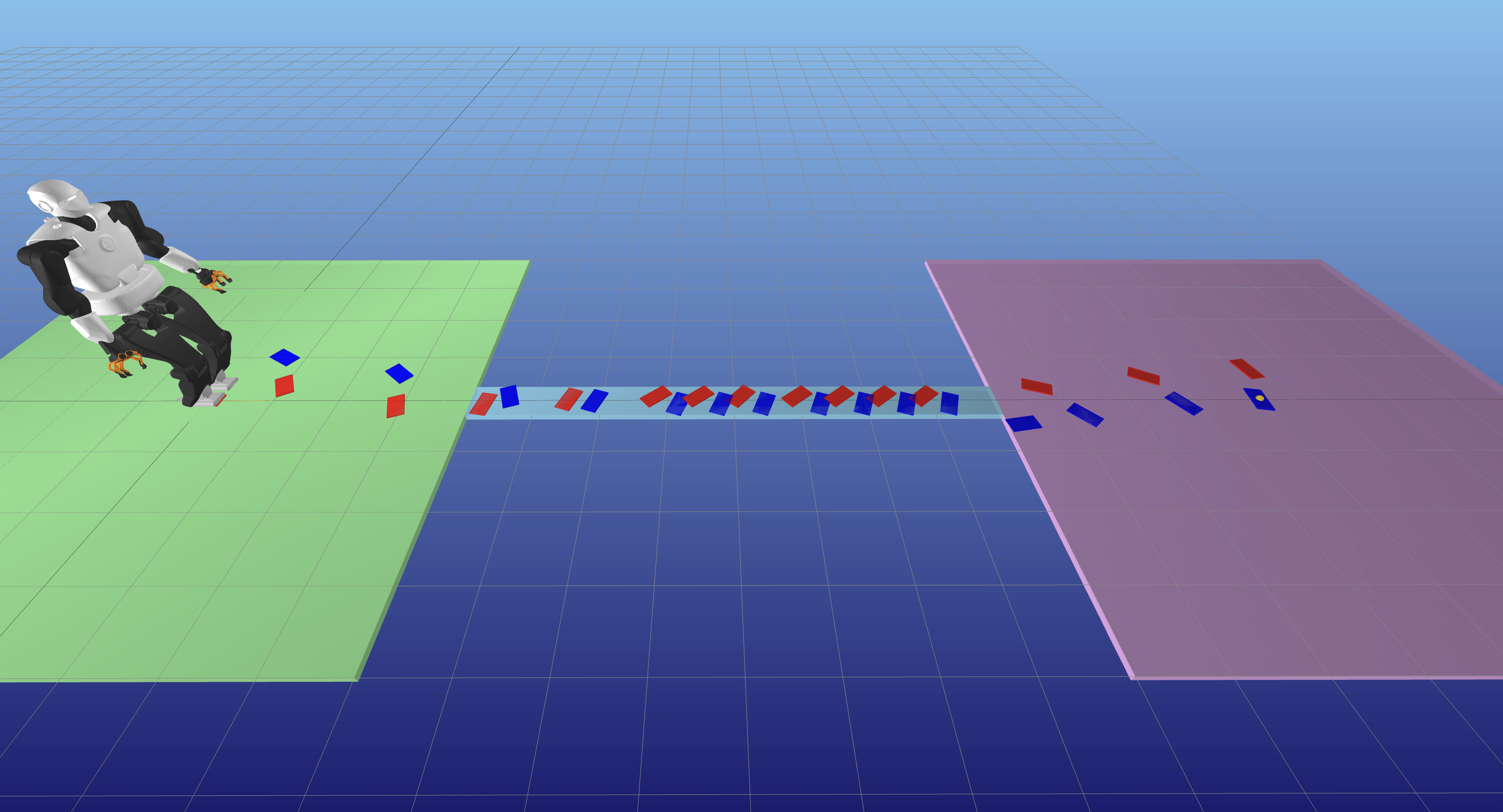}}
    \\
    % ligne 2
    \includegraphics[width=0.3\textwidth]{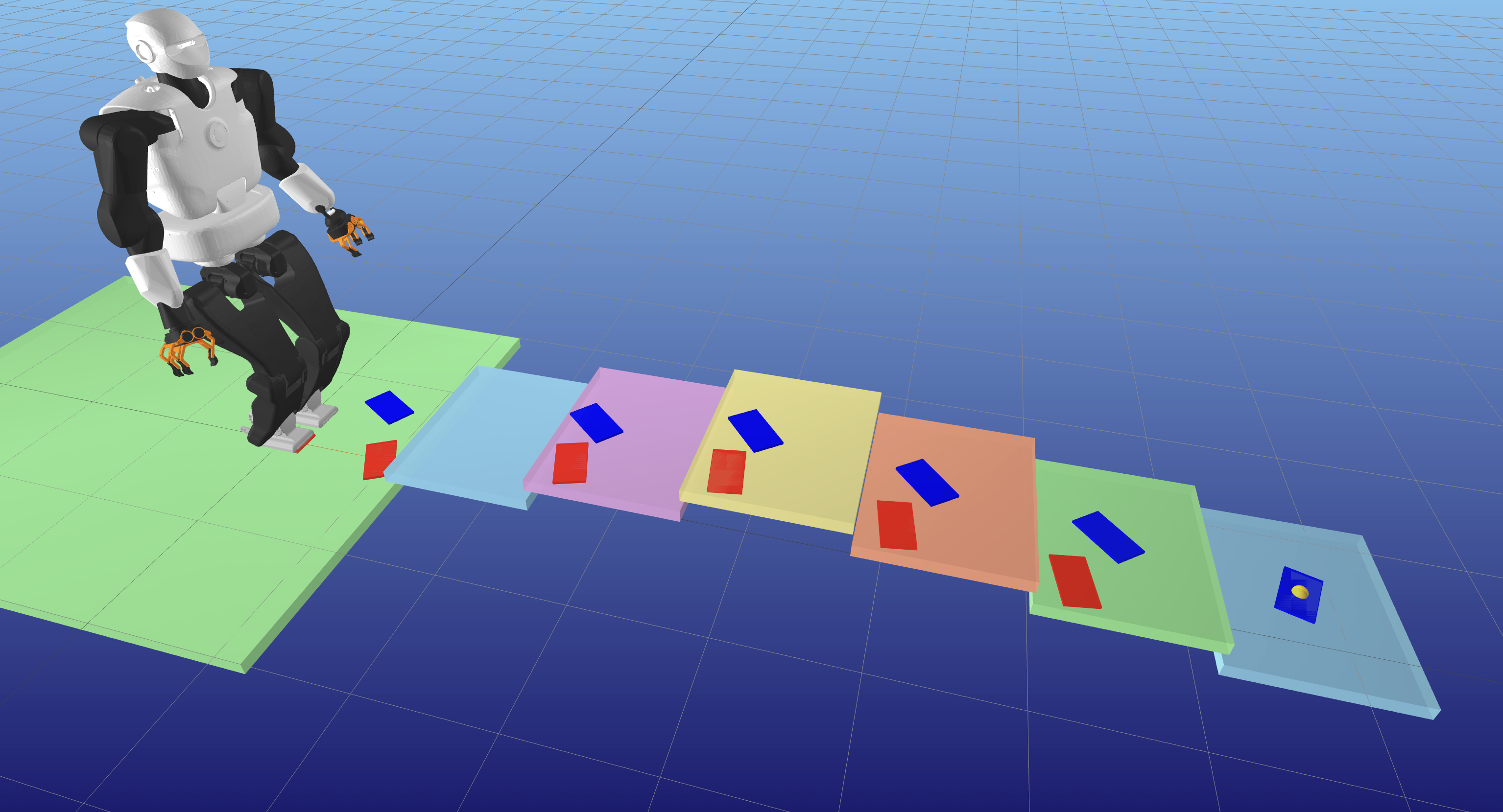}
    \includegraphics[width=0.3\textwidth]{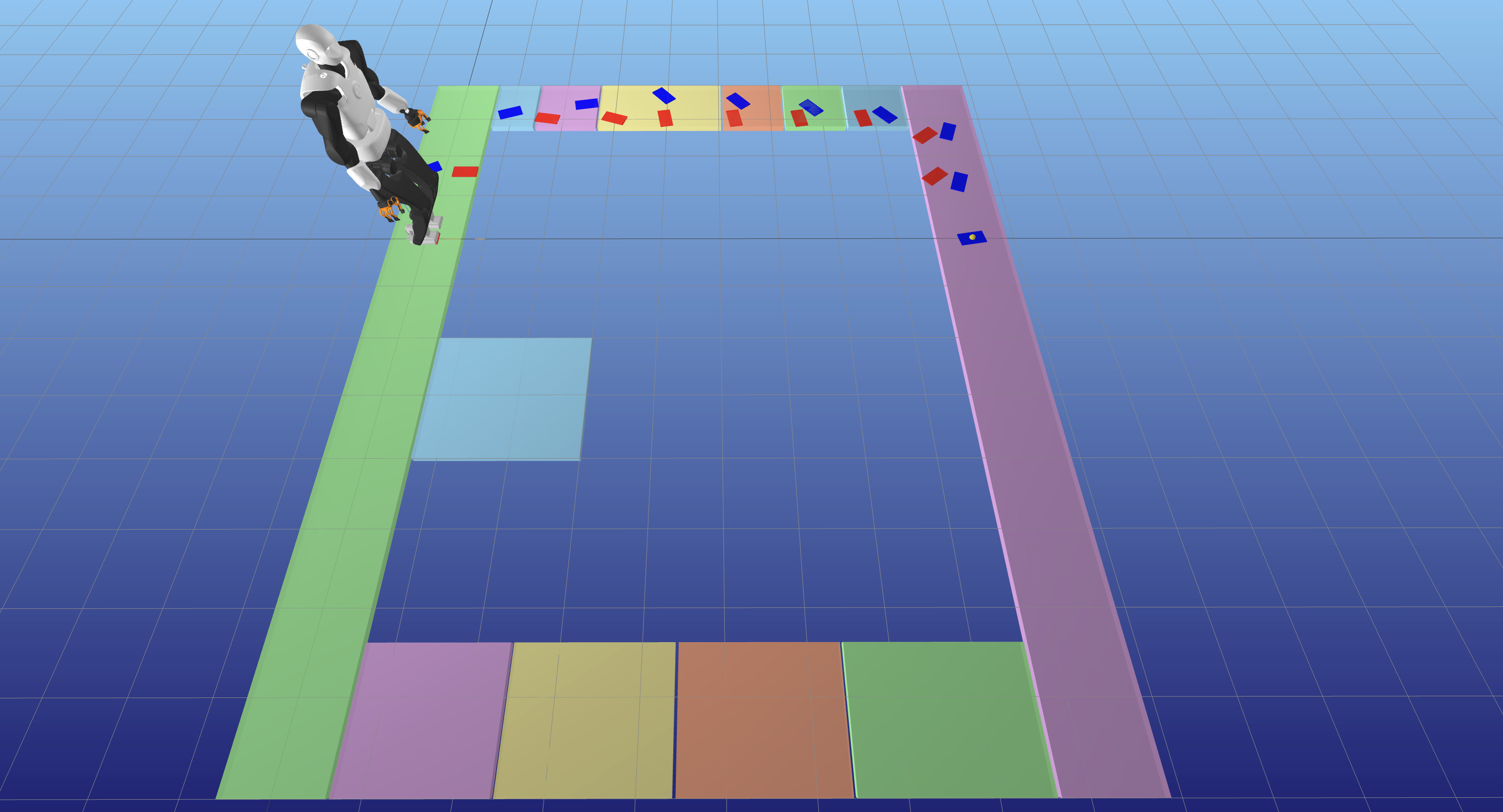}
    \includegraphics[width=0.3\textwidth]{images/s5}
    \caption{Scenarios and contact plans found by \algo{} with (line 2) or without (line 1) rotation. Target is the yellow sphere and Talos is always at the start position. The blue (resp. red) rectangles indicate a contact placement for the left (resp. right).}
    \label{fig:scenarios}
\end{figure*}

We tested three different environments, each designed to test different cases (\reffig{fig:scenarios}). The ``stair'' scenario is an ``easy case'' when the robot only needs to go forward to reach the target, though it needs to select the right surfaces (\reffig{fig:scenarios} - first column). The ``local minima'' is a scenario where a local minima must be escaped before finding a solution (\reffig{fig:scenarios} - second column). The ``narrow passage'' requires the robot to turn sideways to progress, as the reachability constraints prevent stepping forward  (\reffig{fig:scenarios} - third column). In all instances the target is a 3D point for the left foot with no constraints on the final rotation. 
Each scenario is declined into two instances where rotation is enabled or disabled.

\subsection{Results}

% \begin{figure*}[htbp]
%     \centering
%     % première ligne
%     \includegraphics[width=0.4\textwidth]{images/s1}
%     \includegraphics[width=0.4\textwidth]{images/s3}
%     \\
%     % deuxième ligne
%     \includegraphics[width=0.4\textwidth]{images/s2}
%     \includegraphics[width=0.4\textwidth]{images/s4}
%     \\
%     % troisième ligne : seule image
%     \includegraphics[width=0.4\textwidth]{images/s5}
%     \caption{Scenarios and contact plans found by \algo{}. Target is the yellow sphere and Talos is always at the start position. The blue (resp. red) rectangles indicate a contact placement for the left (resp. right) foot}
%      \label{fig:scenarios}
% \end{figure*}

Table \ref{table:result} summarises all results obtained. All experiments were executed 5 times on a computer with Apple M3 CPU (4.05GHz) and 36GB RAM. The presented times correspond to the averages over the runs. In the case of both \astrns, the column \# of Nodes indicates the number of nodes expanded in the graph (not including nodes added but not selected for expansion).
In the case of MIP the column refers to the number of branches explored by the MIP algorithm.

\subsubsection{\algo{} is faster than both MIP and \astrns, although the QP takes time on smaller problems.}
Even when the optimal horizon is given, both \astr formulations outperform the MIP. \algo{} is at least 6 times faster than the optimal MIP, and up to 20 times faster than the 30-step horizon. 

The discretised \astr{}outperforms \algo{} on the easiest instance, even though twice as many nodes are explored by it.  A significant percentage of the computation time of \algo{} is spent solving the QP, which makes the framework less competitive in the easier cases. Using a backward formulation~\cite{nas} would allow to only solve partially the QP without consequences on feasibility. Furthermore, the expansion cost of \algo{} is more expensive. However, as soon as rotation is allowed / the scenarios get more complex, \algo{} is at least twice as fast, and at best more than 100 times faster in the local minima scenario. Of course the performance of the discrete \astr can be improved by lowering the resolution of the discretisation below the value advised in ~\cite{griffin2019footstep}, with consequences on optimality but eventually also on feasibility. With a resolution of 0.1, \astr is still up to 5 times slower than \algo{}  when rotation matters and adds up to 5 steps to the final plan. More importantly, we plan more than 20 steps ahead at 1Hz, enabling real time planning. 

\subsubsection{\algo{} generates less nodes}
The computational gains are explained by the fact that the combinatorics is effectively reduced through the continuous formulation of the constraints. In all scenarios, \algo{} expands fewer nodes than the discretised \astrns, from twice as few to up to 2000 times less in the local minima scenario. 

In the stairs scenario, the number of nodes matches the number of steps (plus the initial node), which implies that no branching occurs at all and the path is immediately found.

The comparison with the MIP solver is not straightforward, as the branching obviously depends on the horizon. Furthermore, the number of nodes explored have a different meaning in both contexts, since a node in \astr represents a specific contact state, while a MIP node represents a branch in the entire search space\footnote{In the stair scenario the 16 nodes explored by \algo{} and the single node explored  the MIP mean that no branching occurred in both cases.}. It is interesting to observe that the pre-solve results in no node expansion, which is in agreement with~\cite{song2020solving}.  However, a horizon of just one extra step in the stairs scenario (i.e. a horizon of 16 steps) results in 36 nodes generated by the solver and reintroduces a combinatorics, worsened if we introduce a cost function.
As soon as the horizon grows, the number of nodes expanded becomes greater than the number of nodes expanded by both \astrns, which is partially linked to a longer path. 

\subsubsection{The minimum number of steps is always found}
Although the EPA heuristic is not admissible, in our experiments the solution found is always optimal with respect to the number of steps (the ground truth was generated using a global search~\cite{nas}). This is only true for the discretised \astrns{} if no rotation is allowed, which can be explained by the loss of possible solutions resulting from the discretisation. 

We also hypothesise that using EPA as cost-to-go provides a better heuristic than the Euclidean distance not only in the rotation case (\cite{griffin2019footstep} indeed suggest that carefully tuned heuristics are usually required to handle rotations), but also when rotation is not allowed, as suggested by the number of node expansions  used in the discrete version. 

\section{discussion}

Our results highlight the advantages of our continuous formulation both in terms of node expansion and computation times, and suggest that our EPA-based heuristic accurately estimates the cost-to-go without requiring cost tuning.

Our two-stage algorithm prevents us from finding the global optimum of a cost function different from the minimal number of steps required, since the choice of contact surface is fixed before the contact locations are optimised. Other cost functions commonly used can only be optimised once the contact sequence is fixed, which is a  limitation of our framework, compared to other works~\cite{cipriano}. This is traded for a better coverage of the feasible space and the guarantee to find a solution under our assumptions. 

 The MIP formulation combines the continuous formulation of \algo{} and the global optimality of a discrete \astrns, but at a significant cost on computational performance. In our context, both \astr algorithms perform orders of magnitude better than the MIP formulation even in the best conditions for the solver, and we hypothesise that when rotations are integrated the combinatorics will explode. Furthermore, it is important to acknowledge the fact that the pre-solve function of gurobi is extremely efficient. Tests with an open source integer or without the pre-solve have resulted in combinatorial explosion and worse results.

As future work, we want to consider the extension of the algorithm to non-cyclic loco-manipulation, but the main challenge remains a better formulation of the reachability constraints that would take into account collisions and dynamics. This issue is common to all approaches compared here, but we believe that as opposed to MIP, \astr based framework could be paired with machine learning to work with more complex, non-linear representations of the constraints.

\section{Conclusion}
We have introduced \algo{}, a novel \astr algorithm based on a continuous formulation of the reachability constraints of a biped robot. Combined with a novel cost-to-go heuristic, we have established experimentally the superiority of the algorithm with respect to both a traditional \astr formulation and a Mixed-Integer program one. 
Our results suggest that long-horizon planning of complex motions could be integrated within a real-time control pipeline for legged robots. 

\balance

\bibliographystyle{IEEEtran}{}
\bibliography{ref}

@string{HUMANOIDS       = "IEEE-RAS Int. Conf. on Hum. Rob."}

@string{IEEE_C_IROS       = "IEEE/RSJ Int. Conf. Intell. Rob. Sys. (IROS)"}

@string{IEEE_C_ICRA       = "IEEE Int. Conf. Rob. Autom. (ICRA)"}

@string{IEEE_J_RAL        = "IEEE Robot. Automat. Lett. (RA-L)"}

@string{IEEE_J_TRO        = "IEEE Trans. Robot. (T-RO)"}

@inproceedings{vandenBergen2001,
  author       = {van den Bergen, Gino},
  title        = {Proximity Queries and Penetration Depth Computation on 3D Game Objects},
  booktitle    = {Proceedings of the Game Developers Conference (GDC)},
  year         = {2001},
  month        = apr,
  address      = {San Jose, California, USA},
  publisher    = {CMP Media},
}

@InProceedings{fukudadouble,
author="Fukuda, Komei
and Prodon, Alain",
editor="Deza, Michel
and Euler, Reinhardt
and Manoussakis, Ioannis",
title="Double description method revisited",
booktitle="Combinatorics and Computer Science",
year="1996",
publisher="Springer Berlin Heidelberg",
address="Berlin, Heidelberg",
pages="91--111",
isbn="978-3-540-70627-4"
}

@article{cipriano,
title = {Humanoid motion generation in a world of stairs},
journal = {Robotics and Autonomous Systems},
volume = {168},
pages = {104495},
year = {2023},
issn = {0921-8890},
doi = {https://doi.org/10.1016/j.robot.2023.104495},
url = {https://www.sciencedirect.com/science/article/pii/S0921889023001343},
author = {Michele Cipriano and Paolo Ferrari and Nicola Scianca and Leonardo Lanari and Giuseppe Oriolo}
}

@article{perrin,
title = "Fast humanoid robot collision-free footstep planning using swept volume approximations",
author = "Nicolas Perrin and Olivier Stasse and L{\'e}o Baudouin and Florent Lamiraux and Eiichi Yoshida",
year = "2012",
month = apr,
doi = "10.1109/TRO.2011.2172152",
language = "English",
volume = "28",
pages = "427--439",
journal = IEEE_J_TRO,
issn = "1552-3098",
publisher = "Institute of Electrical and Electronics Engineers Inc.",
number = "2",

}

@inproceedings{LiuSZ12,
  author       = {Hong Liu and
                  Qing Sun and
                  Tianwei Zhang},
  title        = {Hierarchical {RRT} for humanoid robot footstep planning with multiple
                  constraints in complex environments},
  booktitle    = IEEE_C_IROS,
  pages        = {3187--3194},
  publisher    = {{IEEE}},
  year         = {2012},
  doi          = {10.1109/IROS.2012.6385836},
  bibsource    = {dblp computer science bibliography, https://dblp.org}
}

@ARTICLE{Grandia-perceptive-through-NMPC,
  author={Grandia, Ruben and Jenelten, Fabian and Yang, Shaohui and Farshidian, Farbod and Hutter, Marco},
  journal=IEEE_J_TRO, 
  title={Perceptive Locomotion Through Nonlinear Model-Predictive Control}, 
  year={2023},
  volume={39},
  number={5},
  pages={3402-3421},
  doi={10.1109/TRO.2023.3275384}}

@misc{coalweb,
   author = {Jia Pan and Sachin Chitta and Dinesh Manocha and Florent Lamiraux and Joseph Mirabel and Justin Carpentier and Louis Montaut and others},
   title = {Coal: an extension of the Flexible Collision Library},
   howpublished = {https://github.com/coal-library/coal},
   year = {2015--2024}
}

@INPROCEEDINGS{nas,
  author={Wang, Jiayi and Samadi, Saeid and Wang, Hefan and Fernbach, Pierre and Stasse, Olivier and Vijayakumar, Sethu and Tonneau, Steve},
  booktitle=HUMANOIDS, 
  title={{NAS}: N-step computation of All Solutions to the footstep planning problem}, 
  year={2024},
  volume={},
  number={},
  pages={576-583},
  doi={10.1109/Humanoids58906.2024.10769878}}

@article{althoff2021set,
  title={Set propagation techniques for reachability analysis},
  author={Althoff, Matthias and Frehse, Goran and Girard, Antoine},
  journal={Annual Review of Control, Robotics, and Autonomous Systems},
  volume={4},
  number={1},
  pages={369--395},
  year={2021},
  publisher={Annual Reviews}
}

@inproceedings{bansal2017hamilton,
  title={Hamilton-jacobi reachability: A brief overview and recent advances},
  author={Bansal, Somil and Chen, Mo and Herbert, Sylvia and Tomlin, Claire J},
  booktitle={2017 IEEE 56th Annual Conference on Decision and Control (CDC)},
  pages={2242--2253},
  year={2017},
  organization={IEEE}
}

@article{song2020solving,
  title={Solving Footstep Planning as a Feasibility Problem using L1-norm Minimization (Extended Version)},
  author={Song, Daeun and Fernbach, Pierre and Flayols, Thomas and Del Prete, Andrea and Mansard, Nicolas and Tonneau, Steve and Kim, Young J},
  journal={arXiv preprint arXiv:2011.09772},
  year={2020}
}

@INPROCEEDINGS{wieber,  author={C. {Brasseur} and A. {Sherikov} and C. {Collette} and D. {Dimitrov} and P. {Wieber}},  booktitle={2015 IEEE-RAS 15th International Conference on Humanoid Robots (Humanoids)},   title={A robust linear MPC approach to online generation of 3D biped walking motion},   year={2015},  volume={},  number={},  pages={595-601},  doi={10.1109/HUMANOIDS.2015.7363423}}

@ARTICLE{Winkler,
author={A. W. {Winkler} and C. D. {Bellicoso} and M. {Hutter} and J. {Buchli}},
journal={IEEE Robotics and Automation Letters},
title={Gait and Trajectory Optimization for Legged Systems Through Phase-Based End-Effector Parameterization},
year={2018},
volume={3},
number={3},
pages={1560-1567},
doi={10.1109/LRA.2018.2798285}}

@article{escande2013planning,
  title={Planning contact points for humanoid robots},
  author={Escande, Adrien and Kheddar, Abderrahmane and Miossec, Sylvain},
  journal={Robotics and Autonomous Systems},
  volume={61},
  number={5},
  pages={428--442},
  year={2013},
  publisher={Elsevier}
}

@inproceedings{yunt2006trajectory,
  title={Trajectory optimization of mechanical hybrid systems using sumt},
  author={Yunt, Kerim and Glocker, Christoph},
  booktitle={9th IEEE International Workshop on Advanced Motion Control, 2006.},
  pages={665--671},
  year={2006},
  organization={IEEE}
}

@article{Bretl2006MotionPO,
  title={Motion Planning of Multi-Limbed Robots Subject to Equilibrium Constraints: The Free-Climbing Robot Problem},
  author={T. Bretl},
  journal={The International Journal of Robotics Research},
  year={2006},
  volume={25},
  pages={317 - 342}
}

@MISC{gurobiprimer,
author = {Gurobi},
title = {Mixed-Integer Programming (MIP) – A Primer on the Basics},
howpublished={\url{https://www.gurobi.com/resource/mip-basics}}
}

@article{tonneau2018efficient,
  title={An efficient acyclic contact planner for multiped robots},
  author={Tonneau, Steve and {Del Prete}, Andrea and Pettr{\'e}, Julien and Park, Chonhyon and Manocha, Dinesh and Mansard, Nicolas},
  journal={IEEE Transactions on Robotics},
  volume={34},
  number={3},
  pages={586--601},
  year={2018},
  publisher={IEEE}
}

@inproceedings{stasse2017talos,
  title={TALOS: A new humanoid research platform targeted for industrial applications},
  author={Stasse, Olivier and Flayols, Thomas and Budhiraja, Rohan and Giraud-Esclasse, Kevin and Carpentier, Justin and Mirabel, Joseph and {Del Prete}, Andrea and Sou{\`e}res, Philippe and Mansard, Nicolas and Lamiraux, Florent and others},
  booktitle={2017 IEEE-RAS 17th International Conference on Humanoid Robotics (Humanoids)},
  pages={689--695},
  year={2017},
  organization={IEEE}
}

@inproceedings{griffin2019footstep,
  title={Footstep planning for autonomous walking over rough terrain},
  author={Griffin, Robert J and Wiedebach, Georg and McCrory, Stephen and Bertrand, Sylvain and Lee, Inho and Pratt, Jerry},
  booktitle={2019 IEEE-RAS 19th International Conference on Humanoid Robots (Humanoids)},
  pages={9--16},
  year={2019},
  organization={IEEE}
}

@INPROCEEDINGS{chestnuttkuffner,
  author={Chestnutt, J. and Lau, M. and Cheung, G. and Kuffner, J. and Hodgins, J. and Kanade, T.},
  booktitle=IEEE_C_ICRA, 
  title={Footstep Planning for the Honda ASIMO Humanoid}, 
  year={2005},
  volume={},
  number={},
  pages={629-634},
  doi={10.1109/ROBOT.2005.1570188}}

@inproceedings{deits2014footstep,
  title={Footstep planning on uneven terrain with mixed-integer convex optimization},
  author={Deits, Robin and Tedrake, Russ},
  booktitle={2014 IEEE-RAS international conference on humanoid robots},
  pages={279--286},
  year={2014},
  organization={IEEE}
}

@INPROCEEDINGS{kumagai19,
  author={Kumagai, Iori and Morisawa, Mitsuharu and Benallegue, Mehdi and Kanehiro, Fumio},
  booktitle=HUMANOIDS, 
  title={Bipedal Locomotion Planning for a Humanoid Robot Supported by Arm Contacts Based on Geometrical Feasibility}, 
  year={2019},
  volume={},
  number={},
  pages={132-139},
  doi={10.1109/Humanoids43949.2019.9035072}}

@ARTICLE{kumagailearning,
  author={Kumagai, Iori and Murooka, Masaki and Morisawa, Mitsuharu and Kanehiro, Fumio},
  journal={IEEE Robotics and Automation Letters}, 
  title={Reinforcement Learning of Contact Preferability in Multi-Contact Locomotion Planning for Humanoids}, 
  year={2025},
  volume={10},
  number={2},
  pages={1768-1775},
  doi={10.1109/LRA.2024.3524890}}

@misc{akizhanov2024learningfeasibletransitionsefficient,
      title={Learning feasible transitions for efficient contact planning}, 
      author={Rikhat Akizhanov and Victor Dhédin and Majid Khadiv and Ivan Laptev},
      year={2024},
      eprint={2407.11788},
      archivePrefix={arXiv},
      primaryClass={cs.RO},
      url={https://arxiv.org/abs/2407.11788}, 
}

@ARTICLE{orsolino,
  author={Orsolino, Romeo and Focchi, Michele and Caron, Stéphane and Raiola, Gennaro and Barasuol, Victor and Caldwell, Darwin G. and Semini, Claudio},
  journal={IEEE Transactions on Robotics}, 
  title={Feasible Region: An Actuation-Aware Extension of the Support Region}, 
  year={2020},
  volume={36},
  number={4},
  pages={1239-1255},
  doi={10.1109/TRO.2020.2983318}}

@inproceedings{pettre,
author = {Pettr\'{e}, Julien and Laumond, Jean-Paul and Sim\'{e}on, Thierry},
title = {A 2-stages locomotion planner for digital actors},
year = {2003},
isbn = {1581136595},
publisher = {Eurographics Association},
address = {Goslar, DEU},
booktitle = {Proceedings of the 2003 ACM SIGGRAPH/Eurographics Symposium on Computer Animation},
pages = {258–264},
numpages = {7},
location = {San Diego, California},
series = {SCA '03}
}

@article{Choi03,
author = {Choi, Min Gyu and Lee, Jehee and Shin, Sung Yong},
title = {Planning biped locomotion using motion capture data and probabilistic roadmaps},
year = {2003},
issue_date = {April 2003},
publisher = {Association for Computing Machinery},
address = {New York, NY, USA},
volume = {22},
number = {2},
issn = {0730-0301},
url = {https://doi.org/10.1145/636886.636889},
doi = {10.1145/636886.636889},
journal = {ACM Trans. Graph.},
month = apr,
pages = {182–203},
numpages = {22}
}

@article{kumagai2020multi,
  title={Multi-Contact Locomotion Planning for Humanoid Robot Based on Sustainable Contact Graph With Local Contact Modification},
  author={Kumagai, Iori and Morisawa, Mitsuharu and Hattori, Shizuko and Benallegue, Mehdi and Kanehiro, Fumio},
  journal=IEEE_J_RAL,
  volume={5},
  number={4},
  pages={6379--6387},
  year={2020},
  publisher={IEEE}
}

@Inbook{Hauserprimitive2008,
author="Hauser, Kris
and Bretl, Timothy
and Harada, Kensuke
and Latombe, Jean-Claude",
editor="Akella, Srinivas
and Amato, Nancy M.
and Huang, Wesley H.
and Mishra, Bud",
title="Using Motion Primitives in Probabilistic Sample-Based Planning for Humanoid Robots",
bookTitle="Algorithmic Foundation of Robotics VII: Selected Contributions of the Seventh International Workshop on the Algorithmic Foundations of Robotics",
year="2008",
publisher="Springer Berlin Heidelberg",
address="Berlin, Heidelberg",
pages="507--522",
isbn="978-3-540-68405-3",
doi="10.1007/978-3-540-68405-3_32",
url="https://doi.org/10.1007/978-3-540-68405-3_32"
}

@article{hauser2008motion,
  title={Motion planning for legged robots on varied terrain},
  author={Hauser, Kris and Bretl, Timothy and Latombe, Jean-Claude and Harada, Kensuke and Wilcox, Brian},
  journal={The International Journal of Robotics Research},
  volume={27},
  number={11-12},
  pages={1325--1349},
  year={2008},
  publisher={SAGE Publications Sage UK: London, England}
}

@inproceedings{aceituno2017mixed,
  title={A mixed-integer convex optimization framework for robust multilegged robot locomotion planning over challenging terrain},
  author={Aceituno-Cabezas, Bernardo and Dai, Hongkai and Cappelletto, Jos{\'e} and Grieco, Juan C and Fern{\'a}ndez-L{\'o}pez, Gerardo},
  booktitle={2017 IEEE/RSJ International Conference on Intelligent Robots and Systems (IROS)},
  pages={4467--4472},
  year={2017},
  organization={IEEE}
}

@misc{gurobi,
  author = "Gurobi Optimization, LLC",
  title = "Gurobi Optimizer Reference Manual",
  year = 2019,
  url = "http://www.gurobi.com"
}

@article{piano79,
author = {Lozano-P\'{e}rez, Tom\'{a}s and Wesley, Michael A.},
title = {An algorithm for planning collision-free paths among polyhedral obstacles},
year = {1979},
issue_date = {Oct. 1979},
publisher = {Association for Computing Machinery},
address = {New York, NY, USA},
volume = {22},
number = {10},
issn = {0001-0782},
url = {https://doi.org/10.1145/359156.359164},
doi = {10.1145/359156.359164},
journal = {Commun. ACM},
month = {oct},
pages = {560–570},
numpages = {11}
}

@article{deepgait,
  title={Deepgait: Planning and control of quadrupedal gaits using deep reinforcement learning},
  author={Tsounis, Vassilios and Alge, Mitja and Lee, Joonho and Farshidian, Farbod and Hutter, Marco},
  journal={IEEE Robotics and Automation Letters},
  volume={5},
  number={2},
  pages={3699--3706},
  year={2020},
  publisher={IEEE}
}

@article{deeploco,
  title={Deeploco: Dynamic locomotion skills using hierarchical deep reinforcement learning},
  author={Peng, Xue Bin and Berseth, Glen and Yin, KangKang and Van De Panne, Michiel},
  journal={ACM Trans. on Graph. (TOG)},
  volume={36},
  number={4},
  year={2017},
  publisher={ACM New York, NY, USA}
}

\end{document}